\documentclass{article}
\usepackage{spconf,amsmath,graphicx}
\usepackage{subfig}
\usepackage{multirow}
\usepackage{fancyhdr}

\title{Structural Recurrent Neural Network for Traffic Speed Prediction}
\name{Youngjoo Kim, Peng Wang and Lyudmila Mihaylova \thanks{The authors appreciate the support of the SETA project funded by the European Union’s Horizon 2020 research and innovation program under grant agreement no. 688082.}}
\address{The University of Sheffield\\
		Department of Automatic Control and System Engineering\\
		Sheffield, United Kingdom}
\begin{document}
%

\maketitle


\fancyhf{}
\renewcommand{\headrulewidth}{0pt}
\renewcommand{\footrulewidth}{0pt}
\fancyhead[L]{\textit{Accepted postprint to be presented in ICASSP 2019}}
\fancyfoot[L]{\textcopyright \:2019 IEEE. Personal use of this material is permitted. Permission from IEEE must be obtained for all other uses, in any current or future media, including reprinting/republishing this material for advertising or promotional purposes, creating new collective works, for resale or redistribution to servers or lists, or reuse of any copyrighted component of this work in other works.}%
\thispagestyle{fancy}%

\begin{abstract}

Deep neural networks have recently demonstrated the traffic prediction capability with the time series data obtained by sensors mounted on road segments. However, capturing spatio-temporal features of the traffic data often requires a significant number of parameters to train, increasing computational burden. In this work we demonstrate that embedding topological information of the road network improves the process of learning traffic features. We use a graph of a vehicular road network with recurrent neural networks (RNNs) to infer the interaction between adjacent road segments as well as the temporal dynamics. The topology of the road network is converted into a spatio-temporal graph to form a structural RNN (SRNN). The proposed approach is validated over traffic speed data from the road network of the city of Santander in Spain. The experiment shows that the graph-based method outperforms the state-of-the-art methods based on spatio-temporal images, requiring much fewer parameters to train.

\end{abstract}
\section{Introduction}

Large traffic networks experience a large volume of data and require predictions of the future traffic states based on current and historical traffic data. The traffic data are usually obtained by magnetic induction loop detectors mounted on road segments. These data include traffic speed and flow, where the term traffic flow is used interchangeably with the terms traffic volume and traffic counts. Machine learning approaches have recently been applied to traffic prediction tasks due to the massive volume of traffic data that has become available. The sequence of traffic data on each road segment is essentially a time series. However, each time series pertaining to each road segment has a spatial relationship with each other. Capturing the spatio-temporal patterns of the vehicular traffic is an important task and is part of the control of traffic networks.

Preliminary results on traffic forecasting with convolutional neural networks (CNNs) have been reported \cite{Lv2015, Zhang2017}. They have been demonstrated to be effective in understanding spatial features. Successive convolutional layers followed by max pooling operations increase the field of view of high-level layers and allow them to capture high-order features of the input data. Recurrent neural networks (RNNs) have also been incorporated, considering the traffic prediction as a time series forecasting. Different gating mechanisms like long short-term memories (LSTMs) \cite{Zhang2017, Ma2015} and gated recurrent unit (GRU) \cite{Wu2018} have been tested with various architecture. Instead of dealing with spatial features and temporal features separately, a novel approach has been proposed in \cite{Ma2017} where the traffic data are converted into spatio-temporal images that are fed into a CNN. The deep neural network captures the spatio-temporal characteristics by learning the images. Recently, a capsule network (CapsNet) architecture proposed in \cite{Kim2018} has been demonstrated to outperform the state-of-the-art in complex road networks. The dynamic routing algorithm of the CapsNet replaces the max pooling operation of the CNN, resulting in more accurate predictions but more parameters to train. Gaussian process (GP) \cite{Wang2018} approach is another data-driven approach considered as a kernel-based learning algorithm. GPs have been repeatedly demonstrated to be powerful in exploring the implicit relationship between data to predict the value for an unseen point. Although comparative studies \cite{Xie2010, Chen2015} have shown that GPs are effective in short-term traffic prediction, they still suffer from cubic time complexity in the size of training data.

Inspired by ideas from \cite{Jain2016, Vemula2018}, this paper develops a structural RNN (SRNN) for traffic speed prediction, by incorporating the topological information into the sequence learning capability of RNNs. Considering each road segments as a node, the spatio-temporal relationship is represented by spatial edges and temporal edges. All the nodes and edges are associated with RNNs that are jointly trained. A computationally efficient SRNN is implemented in this paper and the performance is evaluated with real data.

The remaining of the paper is organized as follows. Section 2 describes the traffic speed prediction problem of interest. Section 3 validates the performance of the proposed approach. Finally, Section 4 concludes the paper.

\section{Traffic Speed Prediction}

\subsection{Problem Formulation}

In this study, we address the problem of short-term traffic speed prediction based on historical traffic speed data and a road network graph. Suppose we deal with $N$ road segments where the loop detectors are installed. Let $x_v^t$ represent the traffic speed on road segment $v$ at time step $t$. Given a sequence of traffic speed data $\{x_v^t\}$ for road segments $v = 1, 2, ..., N$ at time steps $t = T-l+1, ..., T$, we predict the future traffic speed $x_v^{T+1}$ on each road segment where $T$ denotes the current time step and $l$ denotes the length of data sequence under consideration.

\subsection{Spatio-Temporal Graph Representation}

We use a spatio-temporal graph representation $\mathcal{G}=(\mathcal{V}, \mathcal{E}_S, \mathcal{E}_T)$ as in \cite{Jain2016, Vemula2018}. Let $\mathcal{G}$ denote the spatio-temporal graph. $\mathcal{V}$, $\mathcal{E}_S$, and $\mathcal{E}_T$ denote the set of nodes, the set of spatial edges, and the set of temporal edges, respectively.  

\begin{figure}[t!]
	\centering
	\subfloat[Spatio-temporal graph]{%
		\includegraphics[clip,width=0.9\linewidth]{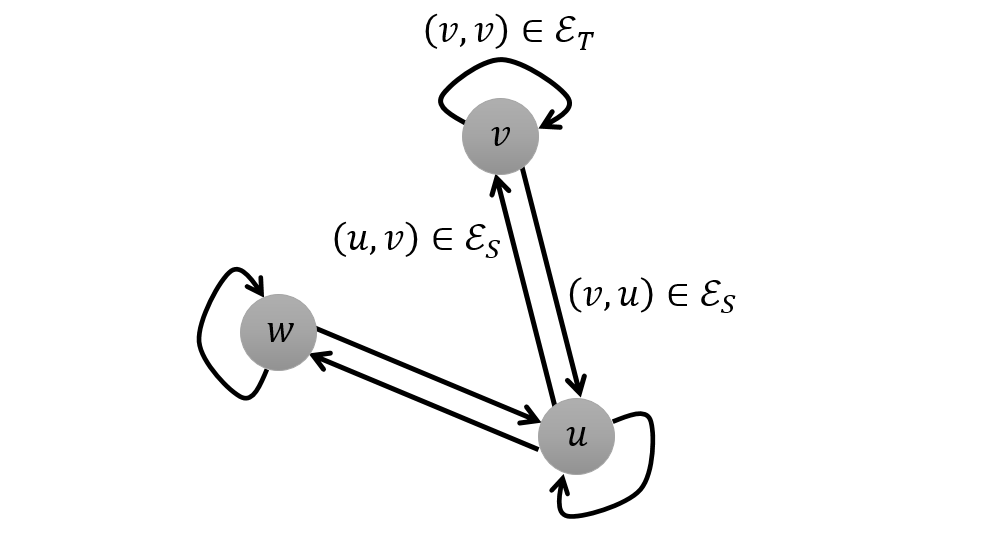}%
	}
	
	\subfloat[Unrolled over time]{%
		\includegraphics[clip,width=0.9\linewidth]{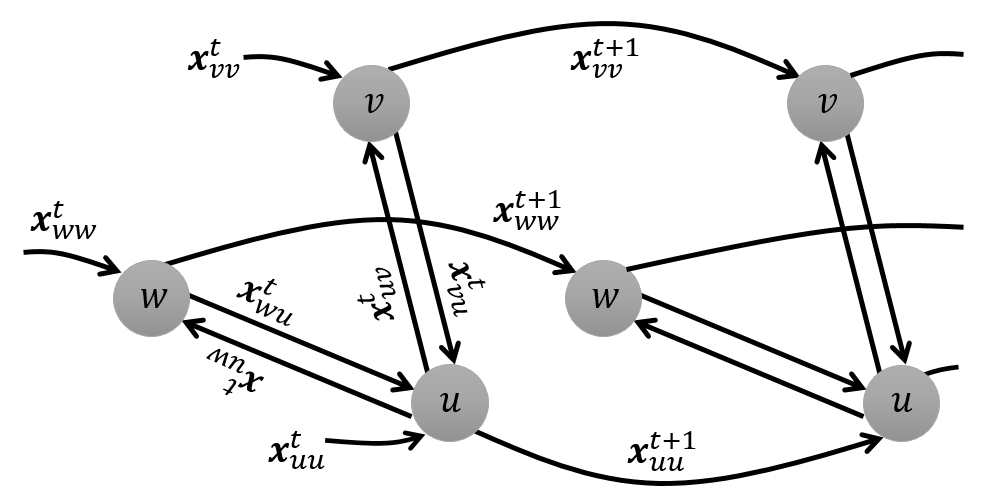}%
	}
	
	\caption{An example spatio-temporal graph. (a) Nodes represent road segments and the nodes are linked by spatial edges $\mathcal{E}_S$ and temporal edges $\mathcal{E}_T$. (b) The spatio-temporal graph is unrolled over time using the temporal edges $\mathcal{E}_T$. The edges are labelled with corresponding feature vectors.}
	\label{fig01}
\end{figure}

In this study, the nodes in the graph correspond to road segments of interest. Thus, $|\mathcal{V}|=N$. The spatial edges represent the dynamics of traffic interaction between two adjacent road segments, and the temporal edges represent the dynamics of the temporal evolution of the traffic speed in road segments. Fig. 1(a) shows an example spatio-temporal graph. Nodes $u, v, w \in \mathcal{V}$ represent road segments. The connection between the road segments is represented by spatial edges $\mathcal{E}_S$. Note that our approach differs from \cite{Vemula2018} in that the spatial edges are established if the two road segments are connected, whereas \cite{Vemula2018} employs an attention model on a fully-connected graph. In addition, we use two spatial edges in opposite direction to link neighbouring nodes. We attempt to take into account the directionality of the interaction between road segments. A temporal edge originated from node $v$ is pointing node $v$. The spatial graph $(\mathcal{V}, \mathcal{E}_S)$ is unrolled over time using temporal edges $\mathcal{E}_T$ to form $\mathcal{G}$ as depicted in Fig. 1(b) where the edges are labelled with corresponding feature vectors.

The feature of node $v \in \mathcal{V}$ at time step $t$ is $x_v^t$, denoting the traffic speed on the road segment. The feature vector of spatial edge $(u,v) \in \mathcal{E}_S$ at time step $t$ is $\mathbf{x}_{uv}^t = [x_u^t, x_v^t] $, which is obtained by concatenating the features of nodes $u$ and $v$ as a row vector. Two spatial edges linking two nodes $u$ and $v$ in opposite direction have different feature vectors, e.g., $\mathbf{x}_{uv}^t = [x_u^t, x_v^t]$ and $\mathbf{x}_{vu}^t = [x_v^t, x_u^t]$. The feature vector of temporal edge $(v,v) \in \mathcal{E}_T$ at time step $t$ is $\mathbf{x}_{vv}^t = [x_v^{t-1}, x_v^t] $, which is obtained by concatenating the features of node $v$ at the previous time step and the current time step.

\subsection{Model Architecture}

\begin{figure}[t!]
	\centering
	\includegraphics[clip,width=0.95\linewidth]{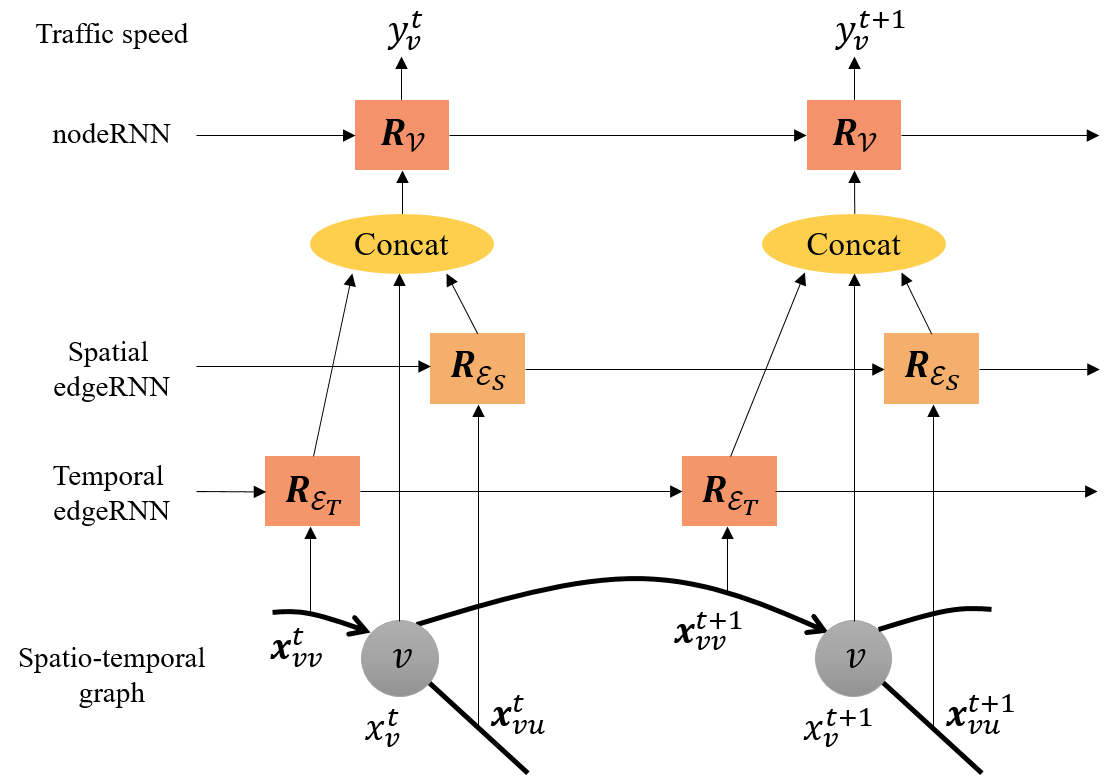}%
	\caption{Architecture of the SRNN in perspective of node $v$ drawn with the unrolled spatio-temporal graph.}
	\label{fig02}
\end{figure}

In our architecture of the SRNN, the sets of nodes $\mathcal{V}$, spatial edges $\mathcal{E}_S$, and temporal edges $\mathcal{E}_T$ are associated with RNNs denoted as nodeRNN $\mathbf{R}_{\mathcal{V}}$, spatial edgeRNN $\mathbf{R}_{\mathcal{E}_S}$, and temporal edgeRNN $\mathbf{R}_{\mathcal{E}_T}$, respectively. The SRNN is derived from the factor graph representation \cite{Jain2016}. Our architecture is the simplest case where the nodes, spatial edges, and temporal edges are sharing the same factors, respectively. This means we assume the dynamics of spatio-temporal interactions is semantically same for all road segments, which keeps the overall parametrization compact and makes the architecture scalable with varying number of road segments. Readers interested in the factor graph representation can refer to \cite{Kschischang2001}.

Fig. 2 visualises the overall architecture. For each node $v$, a sequence of node features $\{x_v^t\}_{t=T-l+1}^T$ is fed into the architecture. Every time each node feature enters, the SRNN is supposed to predict the node label $y_v^{t}$, which corresponds to the traffic speed at the next time step $x_v^{t+1}$. The input into the edgeRNNs is the edge feature $\mathbf{x}_{e}^t$ of edge $e \in \mathcal{E}_S \cup \mathcal{E}_T$ where the edge is incident to node $v$ in the spatio-temporal graph. The node feature $x_v^t$ is concatenated with the outputs of the edgeRNNs to be fed into the nodeRNN. 

We use LSTMs for the RNNs. The hidden state of the nodeRNN has a dimension of 128, and that of the edgeRNNs has a dimension of 256. We employ embedding layers in the network that convert the input into an 128-dimensional vector with a rectified linear unit (ReLU) activation function to give nonlinearity.

\section{Performance Validation}

\subsection{Dataset}

We use a traffic speed dataset from the case studies of the SETA EU project \cite{SETA2016}. Traffic speed measurements had been taken every 15 minutes in the central Santander city of Spain for the year of 2016. Each sparsely missing measurement is masked with an average of speed data recorded at the same time in the other days. We use data of the first 9 months as a training set and the remaining data of the last 3 months as an evaluation set.

We compare the performance of the proposed SRNN with the CapsNet architecture in \cite{Kim2018} that outperforms the state-of-the-art with the dataset. These methods performed the following two speed prediction tasks:
\begin{itemize}
	\item Task 1: prediction based on 10-time-step data ($l=10$)
	\item Task 2: prediction based on 15-time-step data ($l=15$)
\end{itemize}
for two different sets of road segments as depicted in Fig. 3. 50 road segments of interest, where the speed sensors are installed, are marked in red ($N=50$).

\begin{figure}[t!]
	\centering
	\subfloat[]{%
		\includegraphics[clip,width=0.95\linewidth]{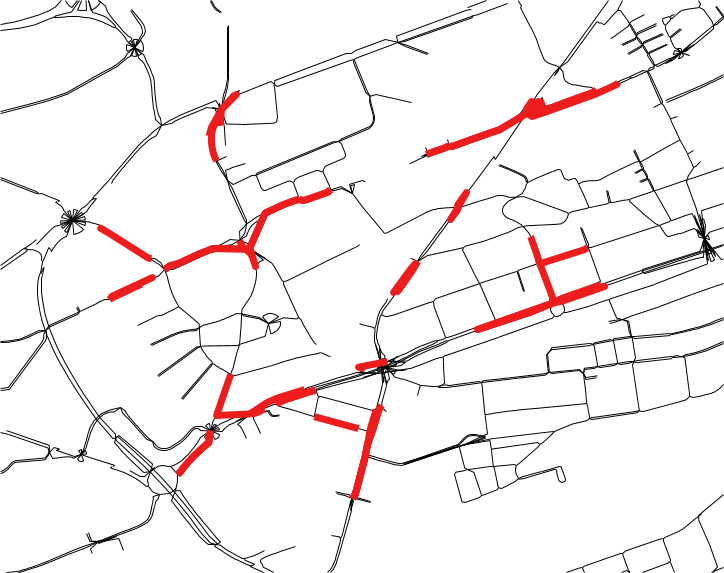}%
	}
	
	\subfloat[]{%
		\includegraphics[clip,width=0.95\linewidth]{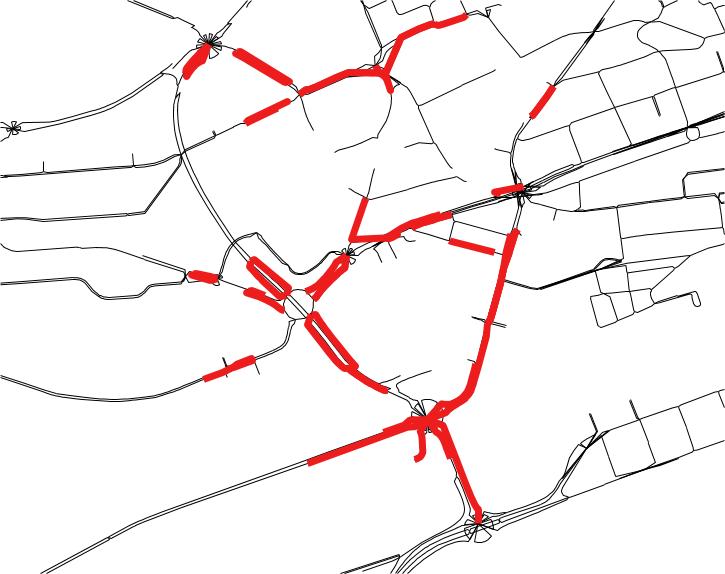}%
	}
	
	\caption{Two sets of road segments used in the experiment. Each set contains 50 road segments marked in red.}
	\label{fig03}
\end{figure}

\subsection{Implementation Details}

As one can see from Fig. 3, the road segments of interest are located sparsely. For constructing the spatial graph in the proposed architecture, we consider road segments adjacent to $v$ if they have the shortest distance to segment $v$. Here, the distance means the number of links traversed from a node to another. Our model has been developed based on the Pytorch implementation of \cite{Vemula2018}. 

The proposed architecture and the CapsNet in \cite{Kim2018} are given their best settings. Our network is trained with a batch size of 8, a starting learning rate of 0.001, and an exponential decay rate of 0.99. The CapsNet is trained with a batch size of 10, a starting learning rate of 0.0005, and an exponential decay rate of 0.9999. Both networks employ the mean squared error (MSE) as a loss function and the Adam optimizer \cite{Kingma2014}. The traffic speed data, measured in [$km/h$], are scaled into the range $[0,1]$ before fed into the networks.

\subsection{Performance Metrics}

Statistical performance metrics are required to validate the overall performance of the networks. The mean relative error (MRE) is one of the most common metrics to quantify the accuracy of different prediction models in general. However, the error of a larger value of speed might result in a smaller MRE and vice versa, providing inconsistent results as witnessed in \cite{Kim2018}.  Thus, we employ mean absolute error (MAE) and root mean squared error (RMSE) as more intuitive metrics for assessing the speed prediction performance. These performance metrics are defined as:

\begin{equation}
MAE = \frac {\sum_{v \in \mathcal{V}, t \in \mathcal{T}} { |y_v^t - \hat{y}_v^t| } } {I} 
\end{equation}
\begin{equation}
RMSE = \sqrt { \frac{\sum_{v \in \mathcal{V}, t \in \mathcal{T}} { (y_v^t - \hat{y}_v^t)^2 }}  {I} }
\end{equation}
where $\hat{y}_v^t$ and $y_v^t$ denote the speed prediction on road segment $v$ at time step $t$ and its true value, respectively. Here, $\mathcal{T}$ denotes the set of time steps in the evaluation set, and
$I=|\mathcal{V}| \times |\mathcal{T}|$ represents the number of the speed data in the evaluation set.

\subsection{Results}

Table 1 shows the resultant performance of the neural networks on the two tasks. The best performance out of 20 epochs is obtained for each method. The result indicates the SRNN performs slightly better by $2.3\%$ in RMSE, showing mere performance difference between methods and between tasks. The distinguishable difference resides in the number of trainable parameters which is translated into computational burden. The number of trainable parameters of the CapsNet varies from $5.1 \times 10^7$ (Task 1) to $7.6 \times 10^7$ (Task 2). Meanwhile, the number of trainable parameters of the SRNN is $1.1 \times 10^6$, which is independent of the sequence length $l$. In fact, the size of the trainable parameter set of the SRNN is affected only by the size of the RNNs. The image-based approaches \cite{Ma2017, Kim2018} would face a significant increase in the computation time as the sequence length $l$ and the number of road segments $N$ increase. On the other hand, the SRNN is scalable to varying $l$ and $N$, and can learn the spatio-temporal traffic characteristic with much fewer parameters given the topological information.

\begin{table}[t!]
	\renewcommand{\arraystretch}{1.5}
	\caption{Speed prediction performance (unit: km/h).}
	\centering
	\begin{tabular}{|l|c|c|c|c|}
		\hline
		\multirow{2}{*}{}                     & \multicolumn{2}{c|}{\textbf{CapsNet}}          & \multicolumn{2}{c|}{\textbf{SRNN}}             \\ \cline{2-5} 
		& \textit{\textbf{MAE}} & \textit{\textbf{RMSE}} & \textit{\textbf{MAE}} & \textit{\textbf{RMSE}} \\ \hline
		\multicolumn{1}{|c|}{\textbf{Task 1}} & 5.720                 & 9.133                  & 5.632                 & 8.906                  \\ \hline
		\multicolumn{1}{|c|}{\textbf{Task 2}} & 5.741                 & 9.172                  & 5.588                 & 8.975                  \\ \hline
	\end{tabular}
	\renewcommand{\arraystretch}{1}
\end{table}

\section{Conclusion}
This paper presents a SRNN architecture that combines the road network map with the traffic speed data to make predictions of future traffic speed. The proposed architecture captures the spatio-temporal relationship of the traffic data with much fewer parameters compared with the image-based state-of-the-art methods.

Existing methods generally provide predictions on road segments where the traffic history is available. Our future work will focus on predictions in road networks with sparse data.


\bibliographystyle{IEEEtran}
\bibliography{ICASSP2019}

\begin{thebibliography}{10}
\providecommand{\url}[1]{#1}
\csname url@samestyle\endcsname
\providecommand{\newblock}{\relax}
\providecommand{\bibinfo}[2]{#2}
\providecommand{\BIBentrySTDinterwordspacing}{\spaceskip=0pt\relax}
\providecommand{\BIBentryALTinterwordstretchfactor}{4}
\providecommand{\BIBentryALTinterwordspacing}{\spaceskip=\fontdimen2\font plus
\BIBentryALTinterwordstretchfactor\fontdimen3\font minus
  \fontdimen4\font\relax}
\providecommand{\BIBforeignlanguage}[2]{{%
\expandafter\ifx\csname l@#1\endcsname\relax
\typeout{** WARNING: IEEEtran.bst: No hyphenation pattern has been}%
\typeout{** loaded for the language `#1'. Using the pattern for}%
\typeout{** the default language instead.}%
\else
\language=\csname l@#1\endcsname
\fi
#2}}
\providecommand{\BIBdecl}{\relax}
\BIBdecl

\bibitem{Lv2015}
Y.~Lv, Y.~Duan, W.~Kang, Z.~Li, and F.-Y. Wang, ``Traffic flow prediction with
  big data: A deep learning approach,'' \emph{IEEE Transactions on Intelligent
  Transportation Systems}, vol.~16, no.~2, pp. 865--873, 2015.

\bibitem{Zhang2017}
J.~Zhang, Y.~Zheng, and D.~Qi, ``Deep spatio-temporal residual networks for
  citywide crowd flows prediction,'' in \emph{Proceedings of the AAAI
  Conference on Aritificial Intelligence}, 2017, pp. 1655--1661.

\bibitem{Ma2015}
X.~Ma, H.~Yu, Y.~Wang, and Y.~Wang, ``Large-scale transportation network
  congestion evolution prediction using deep learning theory,'' \emph{PloS
  one}, vol.~10, no.~3, pp. 1--17, 2015.

\bibitem{Wu2018}
Y.~Wu, H.~Tan, L.~Qin, B.~Ran, and Z.~Jiang, ``A hybrid deep learning based
  traffic flow prediction method and its understanding,'' \emph{Transportation
  Research Part C: Emerging Technologies}, vol.~90, pp. 166--180, 2018.

\bibitem{Ma2017}
X.~Ma, Z.~Dai, Z.~He, J.~Ma, Y.~Wang, and Y.~Wang, ``Learning traffic as
  images: a deep convolutional neural network for large-scale transportation
  network speed prediction,'' \emph{Sensors}, vol.~17, no.~4, p. 818, 2017.

\bibitem{Kim2018}
Y.~Kim, P.~Wang, Y.~Zhu, and L.~Mihaylova, ``A capsule network for traffic
  speed prediction in complex road networks,'' in \emph{Proceedings of the
  Symposium Sensor Data Fusion: Trends, Solutions, and Applications}, 2018,
  Conference Proceedings.

\bibitem{Wang2018}
P.~Wang, Y.~Kim, L.~Vaci, H.~Yang, and L.~Mihaylova, ``Short-term traffic
  prediction with vicinity gaussian process in the presence of missing data,''
  in \emph{Proceedings of the Symposium Sensor Data Fusion: Trends, Solutions,
  and Applications}, 2018, Conference Proceedings.

\bibitem{Xie2010}
Y.~Xie, K.~Zhao, Y.~Sun, and D.~Chen, ``Gaussian processes for short-term
  traffic volume forecasting,'' \emph{Transportation Research Record}, vol.
  2165, no.~1, pp. 69--78, 2010.

\bibitem{Chen2015}
J.~Chen, K.~H. Low, Y.~Yao, and P.~Jaillet, ``Gaussian process decentralized
  data fusion and active sensing for spatiotemporal traffic modeling and
  prediction in mobility-on-demand systems,'' \emph{IEEE Transactions on
  Automation Science and Engineering}, vol.~12, no.~3, pp. 901--921, 2015.

\bibitem{Jain2016}
A.~Jain, A.~R. Zamir, S.~Savarese, and A.~Saxena, ``Structural-rnn: Deep
  learning on spatio-temporal graphs,'' in \emph{Proceedings of the Conference
  on Computer Vision and Pattern Recognition}, 2016, Conference Proceedings,
  pp. 5308--5317.

\bibitem{Vemula2018}
A.~Vemula, K.~Muelling, and J.~Oh, ``Social attention: Modeling attention in
  human crowds,'' in \emph{Proceedings of the International Conference on
  Robotics and Automation}.\hskip 1em plus 0.5em minus 0.4em\relax IEEE, 2018,
  Conference Proceedings, pp. 1--7.

\bibitem{Kschischang2001}
F.~R. Kschischang, B.~J. Frey, and H.-A. Loeliger, ``Factor graphs and the
  sum-product algorithm,'' \emph{IEEE Transactions on information theory},
  vol.~47, no.~2, pp. 498--519, 2001.

\bibitem{SETA2016}
SETA EU Project, A ubiquitous data and service ecosystem for better
  metropolitan mobility, Horizon 2020 Programme, 2016. Available:
  http://setamobility.weebly.com/.

\bibitem{Kingma2014}
D.~P. Kingma and J.~Ba, ``Adam: A method for stochastic optimization,''
  \emph{arXiv preprint arXiv:1412.6980}, 2014.

\end{thebibliography}

\end{document}